%% file: main.tex
\documentclass[runningheads]{llncs}
\usepackage{graphicx}
\usepackage{amsmath,amssymb} 

\usepackage{color}
\usepackage[width=122mm,left=12mm,paperwidth=146mm,height=193mm,top=12mm,paperheight=217mm]{geometry}

\usepackage{epsfig}
\usepackage{graphicx}
\usepackage{amsmath}
\usepackage{amssymb}
\usepackage[pagebackref=true,breaklinks=true,colorlinks]{hyperref}
\newcommand{\reffig}[1]{Figure~\ref{fig:#1}}
\newcommand{\refsec}[1]{Section~\ref{sec:#1}}

\newcommand{\reftbl}[1]{Table~\ref{tbl:#1}}

\newcommand{\refeq}[1]{Equation~\ref{eq:#1}}

\newcommand{\ignorethis}[1]{}

\begin{document}

\pagestyle{headings}
\mainmatter

\title{Generative Visual Manipulation\\
on the Natural Image Manifold}

\titlerunning{Generative Visual Manipulation on the Natural Image Manifold}

\authorrunning{Jun-Yan Zhu, Philipp Kr\"ahenb\"uhl, Eli Shechtman, Alexei A. Efros}

\author{Jun-Yan Zhu\inst{1}, Philipp Kr\"ahenb\"uhl\inst{1}, Eli Shechtman\inst{2},  and Alexei A. Efros\inst{1}}

\institute{University of California, Berkeley\inst{1} \\
Adobe Research\inst{2} \\
}

\maketitle

\begin{abstract}
Realistic image manipulation is challenging because it requires modifying the image appearance in a user-controlled way, while preserving the realism of the result.  Unless the user has considerable artistic skill, it is easy to ``fall off'' the manifold of natural images while editing. In this paper, we propose to learn the natural image manifold directly from data using a generative adversarial neural network. We then define a class of image editing operations, and constrain their output to lie on that learned manifold at all times. The model automatically adjusts the output keeping all edits as realistic as possible. All our manipulations are expressed in terms of constrained optimization and are applied in near-real time. We evaluate our algorithm on the task of realistic photo manipulation of shape and color. The presented method can further be used for changing one image to look like the other, as well as generating novel imagery from scratch based on user's scribbles\footnote{The supplemental video, code, and models are available at our \href{http://people.csail.mit.edu/junyanz/projects/gvm/}{website}.}.

\end{abstract}

\input{intro}
\input{relatedwork}
\input{model}
\input{method}

\input{ui}
\input{implementation}
\input{results}
\input{conclusion}

\clearpage
{\small

\bibliographystyle{splncs}
\bibliography{main}
}

\end{document}

%% file: intro.tex
\section{Introduction}

\begin{figure*}[t]
\begin{center}
 \includegraphics[width=1.0\linewidth]{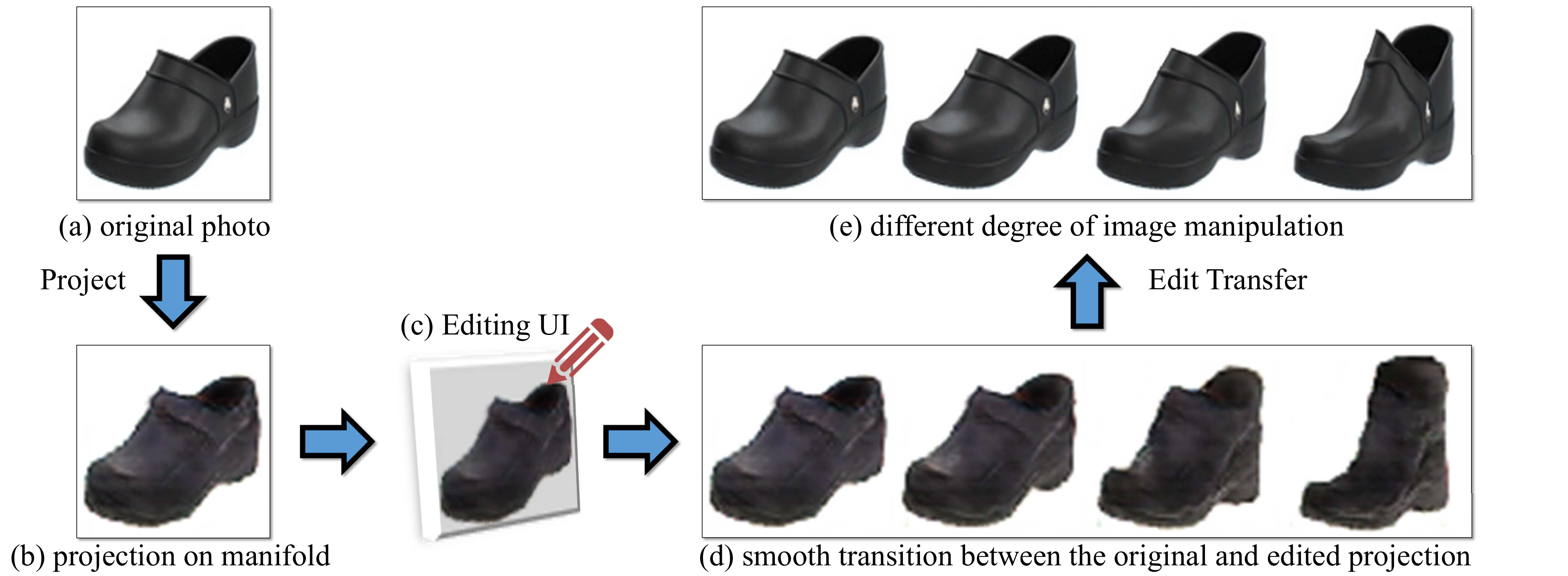}

\end{center}
\vspace{-10pt}
   \caption{We use generative adversarial networks (GAN)~\cite{goodfellow2014generative,radford2015unsupervised} to perform image editing on the natural image manifold. We first project an original photo (a) onto a low-dimensional latent vector representation (b) by regenerating it using GAN.  We then modify the color and shape of the generated image (d) using various brush tools (c) (for example, dragging the top of the shoe). Finally, we apply the same amount of geometric and color changes to the original photo to achieve the final result (e).
   See interactive image editing demo on \href{https://www.youtube.com/watch?v=9c4z6YsBGQ0}{Youtube}. }
\label{fig:teaser}
\vspace{-15pt}
\end{figure*}
Today, visual communication is sadly one-sided. We all perceive information in the visual form (through photographs, paintings, sculpture, etc.), but only a chosen few are talented enough to effectively express themselves visually. This imbalance manifests itself even in the most mundane tasks.  Consider an online shopping scenario: a user looking for shoes has found a pair that mostly suits her but she would like them to be a little taller, or wider, or in a different color.  How can she communicate her preference to the shopping website? If the user is also an artist, then a few minutes with an image editing program will allow her to transform the shoe into what she wants, and then use image-based search to find it.  However, for most of us, even a simple image manipulation in Photoshop presents insurmountable difficulties. One reason is the lack of ``safety wheels'' in image editing: any less-than-perfect edit immediately makes the image look completely unrealistic.  To put another way, classic visual manipulation paradigm does not prevent the user from ``falling off'' the manifold of natural images.

Understanding and modeling the natural image manifold has been a long-standing open research problem.  But in the last two years, there has been rapid advancement, fueled largely by the development of the generative adversarial networks~\cite{goodfellow2014generative}.  In particular, several recent papers~\cite{goodfellow2014generative,radford2015unsupervised,kingma2013auto,denton2015deep,dosovitskiy2016generating} have shown visually impressive results sampling random images drawn from the natural image manifold.   However, two reasons prevent these advances from being useful in practical applications at this time.  First, the generated images, while good, are still not quite photo-realistic (plus there are practical issues in making them high resolution).  Second, these generative models are set up to produce images by sampling a latent vector-space, typically at random.  So, these methods are not able to create and manipulate visual content in a user-controlled fashion.

In this paper, we use the generative adversarial neural network to learn the manifold of natural images, but we do not employ it for image generation.  Instead, we use it as a constraint on the output of various image manipulation operations, to make sure the results lie on the learned manifold at all times.
This idea enables us to reformulate several editing operations, specifically color and shape manipulations, in a natural and data-driven way. The model automatically adjusts the output keeping all edits as realistic as possible (Figure~\ref{fig:teaser}).

We show three applications based on our system: 
(1) Manipulating an existing photo based on an underlying generative model to achieve a different look (shape and color);
(2) ``Generative transformation'' of one image to look more like another;
(3) Generate a new image from scratch based on user's scribbles and warping UI.

All manipulations are performed straightforwardly through gradient-based optimization, resulting in a simple and fast image editing tool. We hope that this work inspires further research in data-driven generative image editing, and thus release the code and data at our \href{http://people.csail.mit.edu/junyanz/projects/gvm/}{website}.

%% file: relatedwork.tex
\section{Prior Work}
\noindent {\bf Image editing and user interaction:} 
Image editing is a well-established area in computer graphics where an input image is manipulated to achieve a certain goal specified by the user. Examples of basic editing include changing the color properties of an image either globally~\cite{reinhard01colortransfer} or locally~\cite{levin2004colorization}.
More advanced editing methods such as image  warping~\cite{alexa2000arap,krahenbuhl2009system} or structured image editing~\cite{barnes2009patchmatch} intelligently reshuffle the pixels in an image following user's edits.
While achieving impressive results in the hands of an expert, when these types of methods fail, they produce results that look nothing like a real image. Common artifacts include unrealistic colors, exaggerated stretching, obvious repetitions, and over-smoothing. This is because they rely on low-level principles (e.g., the similarity of color, gradients or patches) and do not capture higher-level information about natural images.

\vspace{2pt}
\noindent {\bf Image morphing:}  There are several techniques for producing a smooth visual transition between two input images. Traditional morphing methods~\cite{morphingBOOK:Wolberg90} combine an intensity blend with a geometric warp that requires a dense correspondence. In Regenerative Morphing~\cite{shechtman2010regenmorph} the output sequence is regenerated from small patches sampled from the source images. Thus, each frame is constrained to look similar to the two sources. Exploring Photobios~\cite{kemelmacher2011exploring} presented an alternative way to transition between images, by finding the shortest path in a large image collection based on pairwise image distances. Here we extend this idea and produce a morph that is both close to the two sources and stays on, or close to, the natural image manifold.

\vspace{2pt}
\noindent {\bf Natural image statistics:} 
Generative models of local image statistics have long been used as a prior for image restoration problems such as image denoising and deblurring.
A common strategy is to learn local filter or patch models, such as Principal Components, Independent Components, Mixture of Gaussians or wavelet bases~\cite{olshausenField,portillaSimoncelli,zwICCV}. Some methods attempt to capture full-image likelihoods~\cite{rothBlack} through dense patch overlap, though the basic building block is still small patches that do not capture global image structures and long-range relations.
Zhu et al.~\cite{zhu2015learning} recently showed that discriminative deep neural networks learn a much stronger prior that captures both low-level statistics, as well as higher order semantic or color-balance clues. This deep prior can be directly used for a limited set of editing operations (e.g. compositing). However, it does not extend to the diversity of editing operations considered in this work.

\vspace{2pt}
\noindent {\bf Neural generative models:} 
There is a large body of work on neural network based models for image generation. Early classes of probabilistic models of images include restricted Boltzmann machines (e.g., \cite{hinton2006science}) and their deep variants \cite{salakhutdinov2009deepboltzmann},  auto-encoders~\cite{hinton2006science,vincent2008autoencoders} and more recently, stochastic neural networks~\cite{bengio14backprop,kingma2013auto,gregor2015draw} and deterministic networks \cite{dosovitskiy2015learning}.
Generative adversarial networks (GAN), proposed by Goodfellow et al.~\cite{goodfellow2014generative}, learn a generative network jointly with a second discriminative adversarial network in a mini-max objective. The discriminator tries to distinguish between the generated samples and natural image samples, while the generator tries to \textit{fool} the discriminator producing highly realistic looking images. Unfortunately, in practice, GAN does not yield a stable training objective, so several modifications have been proposed recently, such as a multi-scale generation~\cite{denton2015deep} and a convolution-deconvolution architecture with batch normalization~\cite{radford2015unsupervised}.
While the above methods attempt to generate an image starting from a random vector, they do not provide tools to change the generation process with intuitive user controls. In this paper, we remedy this by learning a generative model that can be easily controlled via a few intuitive user edits.

%% file: model.tex
\section{Learning the Natural Image Manifold}

\begin{figure}[t]
\begin{center}
\includegraphics[width=1.0\linewidth]{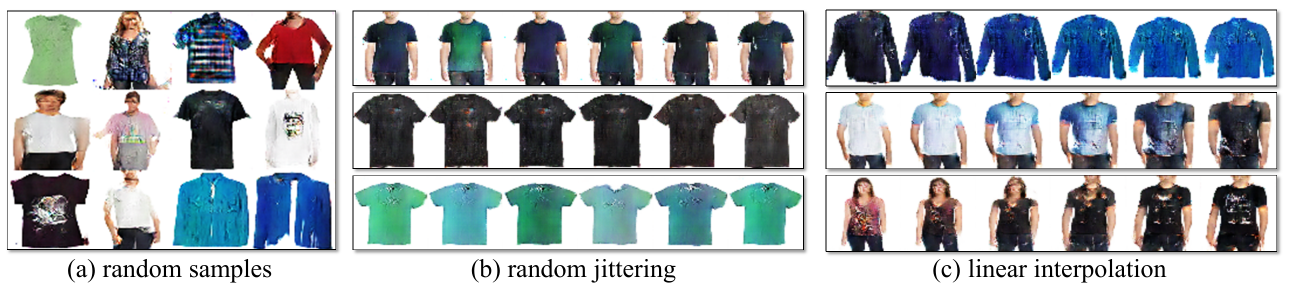}
\end{center}
\vspace{-10pt}
   \caption{GAN as a manifold approximation. (a) Randomly generated examples from a GAN, trained on the shirts dataset; (b) random jittering: each row shows a random sample from a GAN (the first one at the left), and its variants produced by adding Gaussian noise to $z$ in the latent space; (c) interpolation: each row shows two randomly generated images (first and last), and their smooth interpolations in the latent space. }
\label{fig:latent}
\vspace{-15pt}
\end{figure}

Let us assume that all natural images lie on an ideal low-dimensional manifold $\mathbb{M}$ with a distance function $S(x_1, x_2)$ that measures the perceptual similarity between two images $x_1, x_2\in \mathbb{M}$.  Directly modeling this ideal manifold $\mathbb{M}$ is extremely challenging, as it involves training a generative model in a highly structured and complex million dimensional space. Following the recent success of deep generative networks in generating natural looking images, we approximate the image manifold by learning a model using generative adversarial networks (GAN)~\cite{goodfellow2014generative,radford2015unsupervised} from a large-scale image collection. In addition to the high-quality results, GAN has a few other useful properties for our task we will discuss next.

\vspace{2pt}\noindent  {\bf Generative Adversarial Networks:} A GAN model \ignorethis{aims to warp a subspace of d-dimensional space to all possible natural images.  It} consists of two neural networks: (1) a generative network $G(z;\theta_{g})$ that generates an image $x\in \mathbb{R}^{H\times W\times C}$ given a random vector $z\in \mathbb{Z}$, where $\mathbb{Z}$ denotes a d-dimensional latent space, and (2) a discriminative network $D(x;\theta_{d})$ that predicts a probability of a photo being real ($D=1$) or generated ($D=0$). For simplicity, we denote $G(z; \theta_G)$ and $D(x; \theta_D)$ as $G(z)$ and $D(x)$ in later sections. One common choice of $\mathbb{Z}$ is a multivariate uniform distribution $Unif[-1, 1]^d$. $D$ and $G$ are learned using a min-max objective~\cite{goodfellow2014generative}. GAN works well when trained on images of a certain class. We formally define $\tilde {\mathbb M} = \{G(z) | z \in \mathbb{Z}\}$ and use it as an approximation to the ideal manifold ${\mathbb M}$ (i.e $\tilde {\mathbb M} \approx \mathbb{M}$). We also approximate the distance function of two generated images as an Euclidean distance between their corresponding latent vectors, i.e., $S(G(z_1), G(z_2)) \approx \|z_1-z_2\|^2$.

\vspace{2pt}\noindent {\bf GAN as a manifold approximation:}  We use GAN to approximate an ideal manifold for two reasons: first, it produces high-quality samples (see \reffig{latent} (a) for example). Though lacking visual details sometimes, the model can synthesize appealing samples with a plausible overall structure.
Second, the Euclidean distance in the latent space often corresponds to a perceptually meaningful visual similarity (see \reffig{latent} (b) for examples). Therefore, we argue that GAN is a powerful generative model for modeling the image manifold.

\vspace{2pt}\noindent {\bf Traversing the manifold:} Given two images on the manifold $G(z_0), G(z_N)) \in \tilde {\mathbb M}$, one would like to seek a sequence of $N+1$ images $\big[G(z_0), G(z_1), \dots G(z_{N})\big]$ with a smooth transition. This is often done by constructing an image graph with images as nodes, and pairwise distance function  as the edge, and computing a shortest path between the starting image and end image~\cite{kemelmacher2011exploring}. In our case, we minimize $\sum_{t=0}^{N-1} S(G(z_{t}), G(z_{t+1}))$ where $S$ is the distance function. In our case $S(G(z_1), G(z_2))\approx \|z_1-z_2\|^2$ , so a simple linear interpolation $\big[(1-\frac{t}{N}) \cdot z_0 +  \frac{t}{N} \cdot z_{N} \big]_{t=0}^{N}$ is the shortest path. \reffig{latent} (c) shows a smooth and meaningful image sequence generated by interpolating between two points in the latent space. We will now use this approximation of the manifold of natural images for realistic photo editing.

%% file: method.tex
\section{Approach}

\label{sec:method}
\reffig{teaser} illustrates the overview of our approach. Given a real photo, we first project it onto our approximation of the image manifold
by finding the closest latent feature vector $z$ of the GAN to the original image.
Then, we present a real-time method for gradually and smoothly updating the latent vector $z$ so that it generates the desired image that both satisfies the user's edits (e.g., a scribble or a warp; more details in \refsec{ui}) and stays close to the natural image manifold. Unfortunately, in this transformation, the generative model usually looses some of the important low-level details of the input image. Therefore, we propose a dense correspondence method that estimates both per-pixel color and shape changes from the edits applied to the generative model. We then transfer these changes to the original photo using an edge-aware interpolation technique and produce the final manipulated result.

\subsection{Projecting an Image onto the Manifold}
\label{sec:project}

A real photo $x^R$ lies, by definition, on the ideal image manifold $\mathbb{M}$. However for an approximate manifold $\tilde {\mathbb M}$, our goal here is to find a generated image $x^* \in \tilde {\mathbb M}$ close to $x^R$ in some distance metric $\mathcal{L}(x_1, x_2)$ as
\begin{equation} \label{eq:rec_M}
x^* = \underset{x \in \tilde {\mathbb M}}{\arg\min}\ \mathcal{L} (x, \,x^R).
\end{equation}
For the GAN manifold $\tilde {\mathbb M}$ we can rewrite the above equation as follows:
\begin{equation} \label{eq:rec}
z^* = \underset{z \in \tilde {\mathbb Z}}{\arg\min}\ \mathcal{L} (G(z),\, x^R).
\end{equation}
Our goal is to reconstruct the original photo $x^R$ using the generative model $G$ by minimizing the reconstruction error, where $\mathcal{L}(x_1, x_2)=\|\mathcal{C}(x_1)-\mathcal{C}(x_2)\|^2$ in some differentiable feature space $\mathcal{C}$. If $\mathcal{C}(x)=x$, then the reconstruction error is simply pixel-wise Euclidean error. Previous work~\cite{dosovitskiy2016generating,johnson2016perceptual} suggests that using deep neural network activations leads to a reconstruction of perceptually meaningful details. We found that a weighted combination of raw pixels and \textit{conv4} features ($\times 0.002$) extracted from AlexNet~\cite{krizhevsky2012imagenet} trained on ImageNet~\cite{deng2009imagenet} to perform best.

\vspace{2pt}\noindent  {\bf Projection via optimization:}
As both the feature extractor $\mathcal{C}$ and the generative model $G$ are differentiable, we can directly optimize the above objective using L-BFGS-B~\cite{byrd1995limited}. However, the cascade of $\mathcal{C}(G(z))$ makes the problem highly non-convex, and as a result, the reconstruction quality strongly relies on a good initialization of $z$. We can start from multiple random initializations and output the solution with the minimal cost. However, the number of random initializations required to obtain a stable reconstruction is prohibitively large (more than 100), which makes real-time processing impossible.
We instead train a deep neural network to minimize equation \ref{eq:rec} directly.

\vspace{2pt}\noindent  {\bf Projection via a feedforward network:}
We train a feedforward neural network $P(x; \theta_P)$ that directly predicts the latent vector $z$ from a $x$.
The training objective for the predictive model $P$ is written as follows:
\begin{equation} \label{eq:rec_train}
\theta_P^* = \underset{\theta_P}{\arg\min} \sum_{n}  \mathcal{L} (G(P(x_n^R; \theta_P)), \,x_n^R),
\end{equation}
where $x^R_n$ denotes the $n$-th image in the dataset.
The architecture of the model $P$ is equivalent to the discriminator $D$ of the adversarial networks, and only varies in the final number of network outputs.
Objective \ref{eq:rec_train} is reminiscent of an auto-encoder pipeline, with a encoder $P$ and decoder $G$.
However, the decoder $G$ is fixed throughout the training.
While the optimization problem \ref{eq:rec} is the same as the learning objective \ref{eq:rec_train}, the learning-based approach often performs better and does not fall into local optima.
We attribute this behavior to the regularity in the projection problem and the limited capacity of the network $P$. Projections of similar images will share similar network parameters and produce a similar result.
In some sense, the loss for one image provides information for many more images that share a similar appearance~\cite{gershman2014amortized}.
However, the learned inversion is not always perfect, and can often be improved further by a few additional steps of optimization.
\begin{figure}[t]
\begin{center}
 \includegraphics[width=1.0\linewidth]{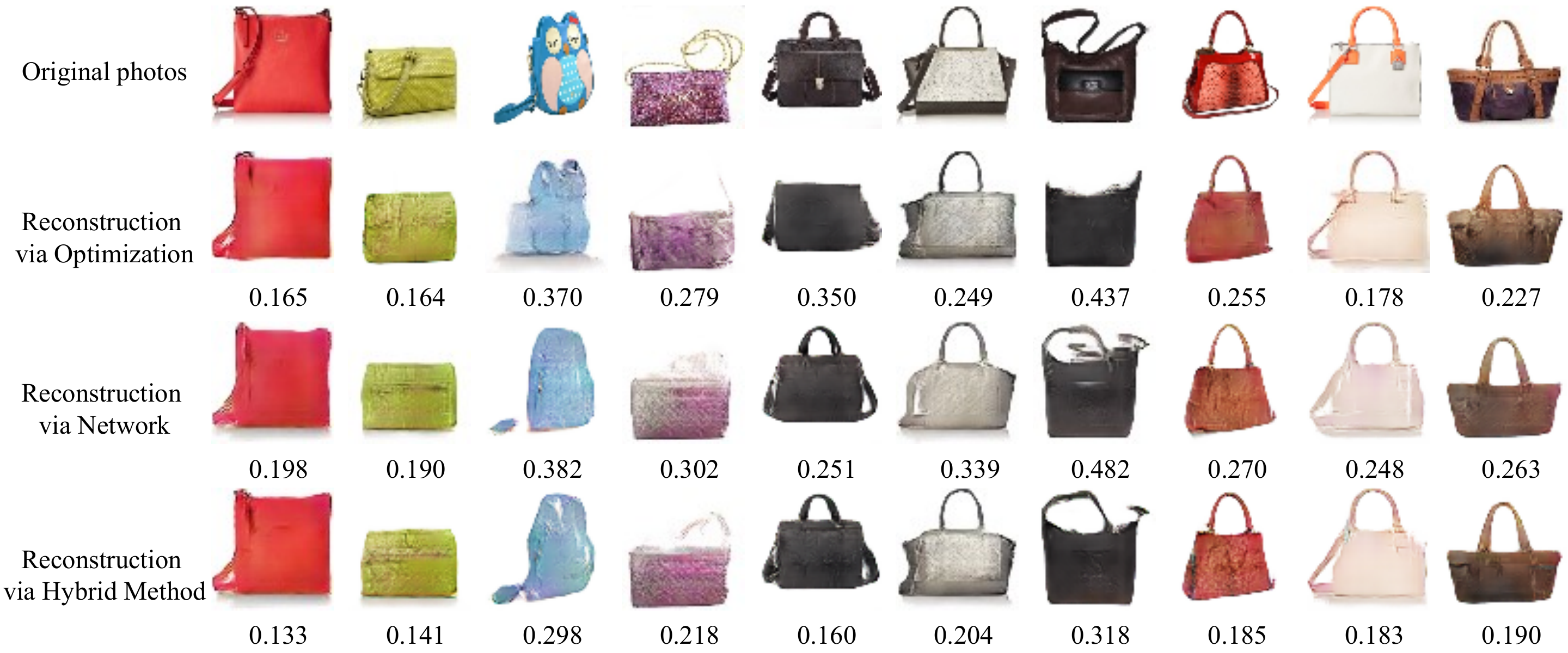}
\end{center}
\vspace{-10pt}
   \caption{Projecting real photos onto the image manifold using GAN. Top row: original photos (from handbag dataset); 2nd row: reconstruction using optimization-based method; 3rd row: reconstruction via learned deep encoder $P$; bottom row: reconstruction using the hybrid method (ours). We show the reconstruction loss below each image.}
\label{fig:rec}
\vspace{-10pt}
\end{figure}

\vspace{2pt}\noindent  {\bf A hybrid method:}
The hybrid method takes advantage of both approaches above. Given a real photo $x^R$, we first predict $P(x^R; \theta_P)$ and then use it as the initialization for the optimization objective (\refeq{rec}). So the learned predictive model serves as a fast bottom-up initialization method for a non-convex optimization problem. ~\reffig{rec} shows a comparison of these three methods. See \refsec{eva} for a more quantitative evaluation.

\subsection{Manipulating the Latent Vector}
\label{sec:update}

\begin{figure}[t]
\begin{center}
 \includegraphics[width=1.0\linewidth]{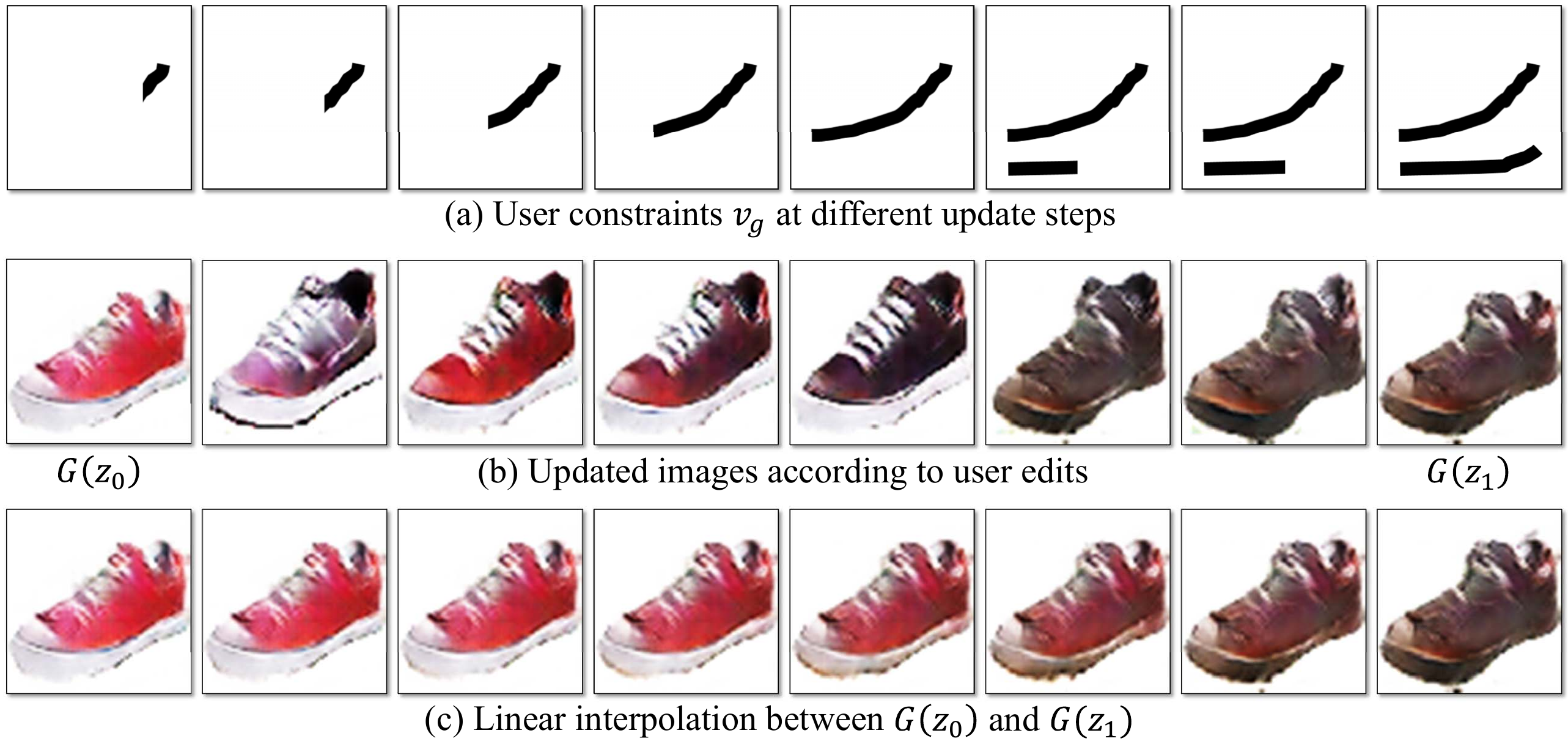}
\end{center}
\vspace{-10pt}
   \caption{Updating latent vector given user edits. (a) Evolving user constraint $v_g$ (black color strokes) at each update step; (b) intermediate results at each update step ($G(z_0)$ at leftmost, and $G(z_1)$ at rightmost); (c) a smooth linear interpolation in latent space between $G(z_0)$ and $G(z_1)$. }
\label{fig:update}
\vspace{-10pt}
\end{figure}

With the image $x_0^R$ projected onto the manifold $\tilde{\mathbb M}$ as $x_0=G(z_0)$ via the projection methods just described, we can start modifying the image on that manifold.
We update the initial projection $x_0$ by simultaneously matching the user intentions while staying on the manifold, close to the original image $x_0$.

Each editing operation is formulated as a constraint $f_g(x) = v_g$ on a local part of the output image $x$.
The editing operations $g$ include color, shape and warping constraints, and are further described in \refsec{brush}.
Given an initial projection $x_0$, we find a new image $x \in \mathbb{M}$ close to $x_0$ trying to satisfy as many constraints as possible
\begin{equation} \label{eq:obj_M}
x^* = \underset{x \in \mathbb{M}}{\arg\min} \Big \{ \underbrace{\sum_g \|f_g(x)-v_g\|^2}_{\text{data term}}+ \underbrace{\lambda_s \cdot S(x, x_0)}_{\begin{subarray}{c}\text{manifold} \\ \text{smoothness} \end{subarray}}\Big \},
\end{equation}
where the data term measures deviation from the constraint and the smoothness term enforces moving in small steps on the manifold, so that the image content is not altered too much. We set $\lambda_s=5$ in our experiments.

The above equation simplifies to the following on the approximate GAN manifold $\tilde {\mathbb M}$:
\begin{equation} \label{eq:obj}
z^* = \underset{z \in \mathbb{Z}}{\arg\min} \Big \{ \underbrace{\sum_g \|f_g(G(z))-v_g\|^2}_{\text{data term}}+ \underbrace{\lambda_s \cdot \|z-z_0\|^2}_{\begin{subarray}{c}\text{manifold} \\ \text{smoothness} \end{subarray}} + E_D\Big \}.
\end{equation}
Here the last term $E_D=\lambda_D \cdot \log (1-D(G(z)))$ optionally captures the visual realism of the generated output as judged by the GAN discriminator $D$.
This constraint further pushes the image towards the manifold of natural images and slightly improves the visual quality of the result.
By default, we turn off this term to increase frame rates.

\vspace{2pt}\noindent  {\bf  Gradient descent update:} For most constraints \refeq{obj} is non-convex. We solve it using gradient descent, which allows us to provide the user with a real-time feedback as she manipulates the image.
As a result, the objective \ref{eq:obj} evolves in real-time as well.
For computational reasons, we only perform a few gradient descent updates after changing the constraints $v_g$.
Each update step takes $50 - 100$ ms, which ensures interactive feedback.
\reffig{update} shows one example of the update of  $z$. Given an initial red shoe as shown in \reffig{update}, the user gradually scribbles a black color stroke (i.e., specifies a region is black) on the shoe image (\reffig{update} a). Then our update method smoothly changes the image appearance (\reffig{update} b) by adding more and more of the user constraints.
Once the final result $G(z_1)$ is computed, a user can see the interpolation sequence between the initial point $z_0$ and $z_1$ (\reffig{update} c), and select any intermediate result as the new starting point. Please see the supplemental video for more details.

While this editing framework allows us to modify any generated image on the approximate natural image manifold $\tilde{\mathbb M}$, it does not directly provide us with a way to alter the original high-resolution image $x_0^R$. In the next section, we show how edits on the approximate manifold can be transferred to the original image.

\subsection{Edit Transfer}
\label{sec:flow}

\begin{figure}[t]
\begin{center}
 \includegraphics[width=1.0\linewidth]{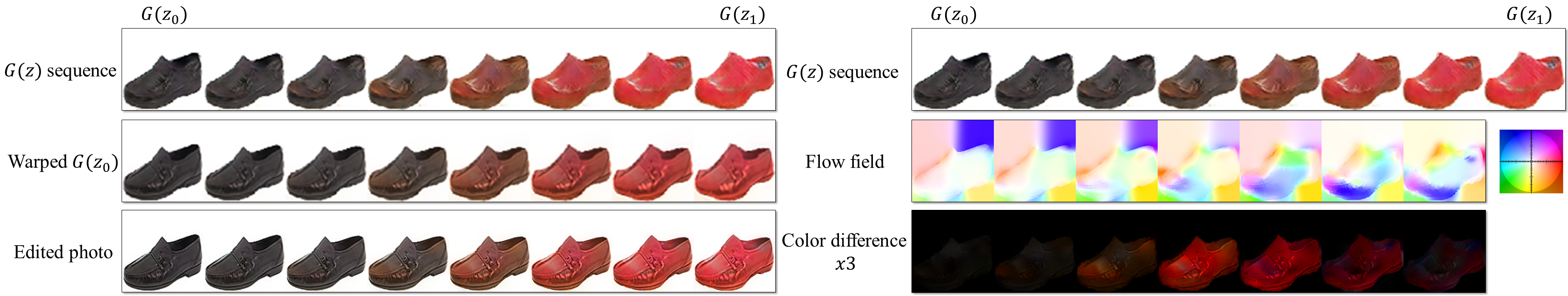}
\end{center}
   \caption{Edit transfer via Motion+Color Flow.
   Following user edits on the left shoe $G(z_0)$ we obtain an interpolation sequence in the generated latent space $G(z)$ (top right).
   We then compute the motion and color flows (right middle and bottom) between neighboring images in $G(z)$.
   These flows are concatenated and, as a validation, can be applied on $G(z_0)$  to obtain a close reconstruction of $G(z)$ (left middle). The bottom left row shows how the edit is transferred to the original shoe using the same concatenated flow, to obtain a sequence of edited shoes.}
\label{fig:flow}
\end{figure}

Give the original photo $x^R_0$ (e.g., a black shoe) and its projection on the manifold $G(z_0)$, and a user modification $G(z_1)$ by our method (e.g., the generated red shoe). The generated image $G(z_1)$ captures the roughly change we want, albeit the quality is degraded w.r.t the original image.

Can we instead adjust the original photo and produce a more photo-realistic result $x^R_1$ that exhibits the changes in the generated image?
A straightforward way is to transfer directly the pixel changes (i.e., $x^R_1=x^R_0+(G(z_1)-G(z_0))$.
We tried this approach, and it introduced new artifacts due to the misalignment of the two images. To address this issue, we develop a dense correspondence algorithm to estimate both the geometric and color changes induced by the editing process.

Specifically, given two generated images $G(z_0)$ and $G(z_1)$, we can generate any number of intermediate frames $\big[G((1-\frac{t}{N}) \cdot z_0 +  \frac{t}{N} \cdot z_1) \big]_{t=0}^N$, where consecutive frames only exhibit minor visual variations.

\vspace{2pt}\noindent  {\bf Motion+Color flow algorithm:} We then estimate the color and geometric changes by generalizing the brightness constancy assumption in traditional optical flow methods~\cite{brox2004high,bruhn2005lucas}.
This results in the following motion+color flow objective\footnote{For simplicity, we omit the pixel subscript $(x,y)$ for all the variables.}:

\begin{equation} \label{eq:flow}
\iint\!\underbrace{\| I\!(x,y,t)\!-\!A\!\cdot\!I\!(x\!+\!u, y\!+\!v, t\!+\!1)\! \|^2\!}_{\text{data term}}+\underbrace{\sigma_s (\|\nabla\!u\|^2\!\!+\!\|\nabla\!v\|^2) \!}_{\text{spatial reg}} + \underbrace {\sigma_c  \|\nabla\!A\|^2\!}_{\text{color reg}}\!dxdy,
\end{equation}
where $I(x,y,t)$ denotes the RGB values $(r,g,b,1)^T$ of pixel $(x,y)$ in the generated image $G((1-\frac{t}{N}) \cdot z_0 +  \frac{t}{N} \cdot z_1)$.
$(u,v)$ is the flow vector with respect to the change of $t$, and $A$ denotes a $3\times4$ color affine transformation matrix. The data term relaxes the color constancy assumption by introducing a locally affine color transfer model $A$~\cite{shih2013data} while the spatial and color regularization terms encourage smoothness in both the motion and color change.
We solve the objective by iteratively estimating the flow $(u,v)$ using a traditional optical flow algorithm, and computing the color change $A$ by solving a system of linear equations~\cite{shih2013data}. We iterate $3$ times. We produce $8$ intermediate frames (i.e., $N=7$).

We estimate the changes between nearby frames, and concatenate these changes frame by frame to obtain long-range changes between any two frames along the interpolation sequence $z_0 \rightarrow z_1$. \reffig{flow} shows a warping sequence after we apply the flow to the initial projection $G(z_0)$.

\vspace{2pt}\noindent  {\bf Transfer edits to the original photo:}
After estimating the color and shape changes in the generated image sequence, we apply them to the original photo and produce an interesting transition sequence of photo-realistic images as shown in \reffig{flow}.
As the resolution of the flow and color fields are limited to the resolution of the generated image ($64\times64$), we upsample those edits using a guided image filter~\cite{he2010guided}.

%% file: ui.tex
\section{User Interface}
\label{sec:ui}
The user interface consists of the main window showing the current edited photo, a display showing thumbnails of all the candidate results, and a slider bar to explore the interpolation sequence between the original photo and the final result. Please see our supplemental video for more details.

\vspace{2pt}\noindent {\bf Candidate results:} 
Given the objective (\refeq{obj}) derived with the user guidance, we generate multiple different results by initializing $z$ as random perturbations of $z_0$. We generate $64$ examples and show the best $9$ results sorted by the objective cost (\refeq{obj}).

\vspace{2pt}\noindent  {\bf Relative edits:} 
Once a user finishes one edit, she can drag a slider to see all the intermediate results interpolated between the original and the final manipulated photo. We call this ``relative edits'' as it allows a user to explore more alternatives with a single edit. Similar to relative attributes~\cite{parikh2011relative}, a user can express ideas like changing the handle of the handbag to be more red, or making the heel of the shoes slightly higher, without committing to a specific final state.

\subsection{Editing constraints}
\label{sec:brush}
Our system provides three constraints for editing the photo in different aspects: coloring, sketching and warping. We express all constraints as brush tools. In the following, we explain the usage of each brush and the corresponding constraints.

\vspace{2pt}\noindent  {\bf Coloring brush:} 
The coloring brush allows the user to change the color of a specific region. The user selects a color from a palette and can adjust the brush size. For each pixel marked with this brush we constrain the color $f_g(I) = I_p = v_g$ of a pixel $p$ to the selected values $v_g$.

\vspace{2pt}\noindent  {\bf Sketching brush:} 
The sketching brush allows the user to outline the shape or add fine details.
We constrain $f_g(I) = HOG(I)_p$ a differentiable HOG descriptor~\cite{dalal2005histograms} at a certain location $p$ in the image to be close to the user stroke (i.e. $v_g=HOG(stroke)_p$). We chose the HOG feature extractor because it is binned, which makes it robust to sketching inaccuracies.

\vspace{2pt}\noindent  {\bf Warping brush:} 
The warping brush allows the user to modify the shape more explicitly. The user first selects a local region (a window with adjustable size), and then drag it to another location. We then place both a color and sketching constraint on the displaced pixels encouraging the target patch to mimic the appearance of the dragged region.

\reffig{sketch2gen} shows a few examples where we use the coloring and sketching brushed for interactive image generation.  \reffig{teaser} shows the result of the warping brush that was used to pull the top line of the shoe up. \reffig{edit} shows a few more examples. 

%% file: implementation.tex
\section{Implementation Details}
\noindent {\bf Network architecture:} We follow the same architecture of deep convolutional generative adversarial networks (DCGAN)~\cite{radford2015unsupervised}.
DCGAN mainly builds on multiple convolution, deconvolution and ReLU layers, and eases the min-max training via batch normalization~\cite{ioffe2015batch}. We train the generator $G$ to produce a $64\times64\times3$ image given a $100$-dimensional random vector. Notice that our method can also use other generative models (e.g. variational auto-encoder~\cite{kingma2013auto} or future improvements in this area) to approximate the natural image manifold.

\noindent {\bf Computational time:} We run our system on a Titan X GPU. Each update of the vector $z$ takes $50\sim 100$ milliseconds, which allows the real-time image editing and generation. Once an edit is finished, it takes  $5\sim10$ seconds for our edit transfer method to produce high-resolution final result. 

%% file: results.tex
\section{Results}
We first introduce the statistics of our dataset. We then show three main applications: realistic image manipulation, generative image transformation, and generating a photo from scratch using our brush tools. Finally, we evaluate our image reconstruction methods and perform a human perception study to understand the realism of generated results. Please refer to the supplementary material for more results and comparisons.

\vspace{2pt}\noindent {\bf Datasets:} \ We experiment with multiple photo collections from various sources as follows: ``shoes'' dataset~\cite{yu2014fine}, which has $50$K shoes collected from Zappos.com (the shoes are roughly centered, but not well aligned, and roughly facing left, with frontal to side view); ``church outdoor'' dataset ($126$K images) from the LSUN challenge~\cite{yu2015construction}; ``outdoor natural'' images ($150$K) from the MIT Places dataset~\cite{zhou2014learning}; and two query-based product collections downloaded from Amazon, including ``handbags'' ($138$K) and ``shirts'' ($137$K). The downloaded handbags and shirts are roughly centered but no further alignment has been performed.

\subsection{Image Manipulation}
\begin{figure}[t]
\begin{center}
\includegraphics[width=0.95\linewidth]{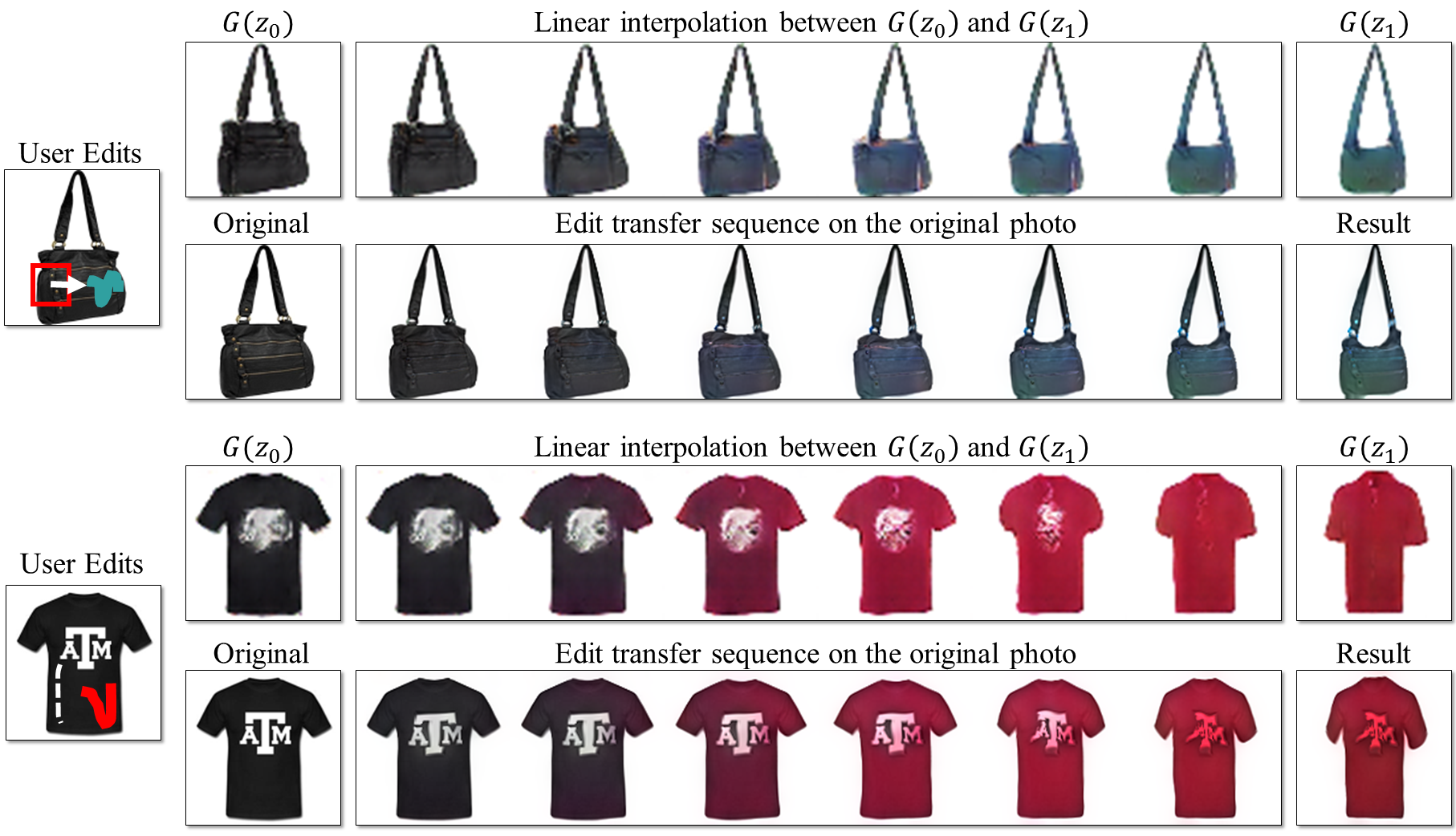}
\end{center}
   \caption{Image manipulation examples: for each example, we show the original photo and user edits on the left. The top row on the right shows the generated sequence and the bottom row shows the edit transfer sequence on the original image.}
\label{fig:edit}
\end{figure}
\label{sec:results}

Our main application is photo-realistic image manipulation using the brush interactions described in \refsec{brush}. See \reffig{edit} for a few examples where the brush edits are depicted on the left (dashed line for the sketch tool, color scribble for the color brush and a red square and an arrow for the warp tool).
See the supplementary video for more interactive manipulation demos.

\subsection{Generative Image Transformation}
\begin{figure}
\begin{center}
\includegraphics[width=1\linewidth]{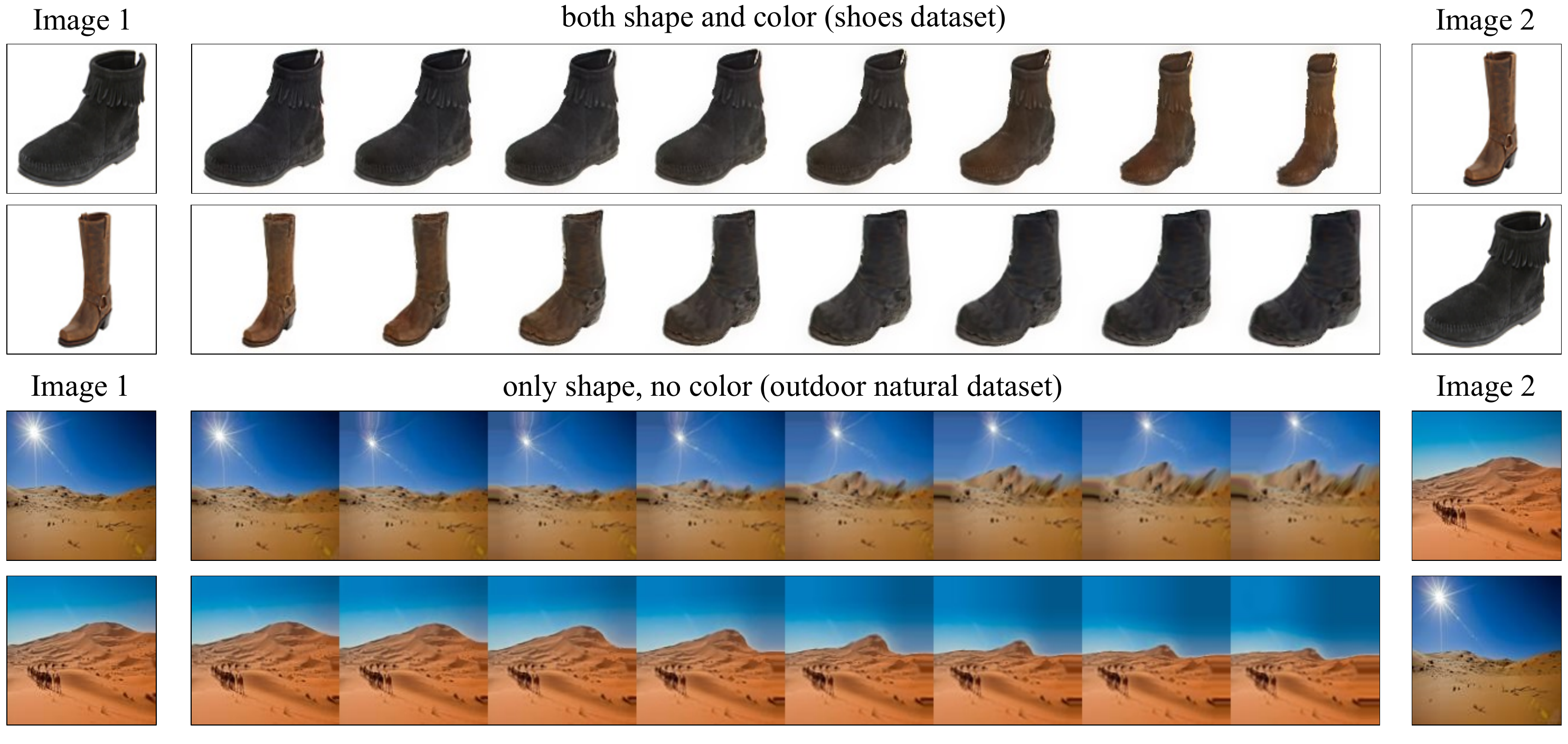}
\end{center}
   \caption{Generative image transformation. In both rows, the source on the left is transformed to have the shape and color (or just shape in the 2nd example) of the one on the right. }
\label{fig:gentransform}
\end{figure}

An interesting outcome of the editing process is the sequence of intermediate generated images that can be seen as a new kind of image morphing~\cite{morphingBOOK:Wolberg90,ViewMorphing:SeitzDyer96,shechtman2010regenmorph}. We call it ``generative transformation''.
We use this sequence to transform the shape and color of one image to look like another image automatically, i.e., {\em without} any user edits. This manipulation is done by applying the motion+color flow  on either of the sources. \reffig{gentransform} shows a few ``generative transform'' examples.

\subsection{Interactive Image Generation}

\begin{figure}
\begin{center}
\includegraphics[width=1\linewidth]{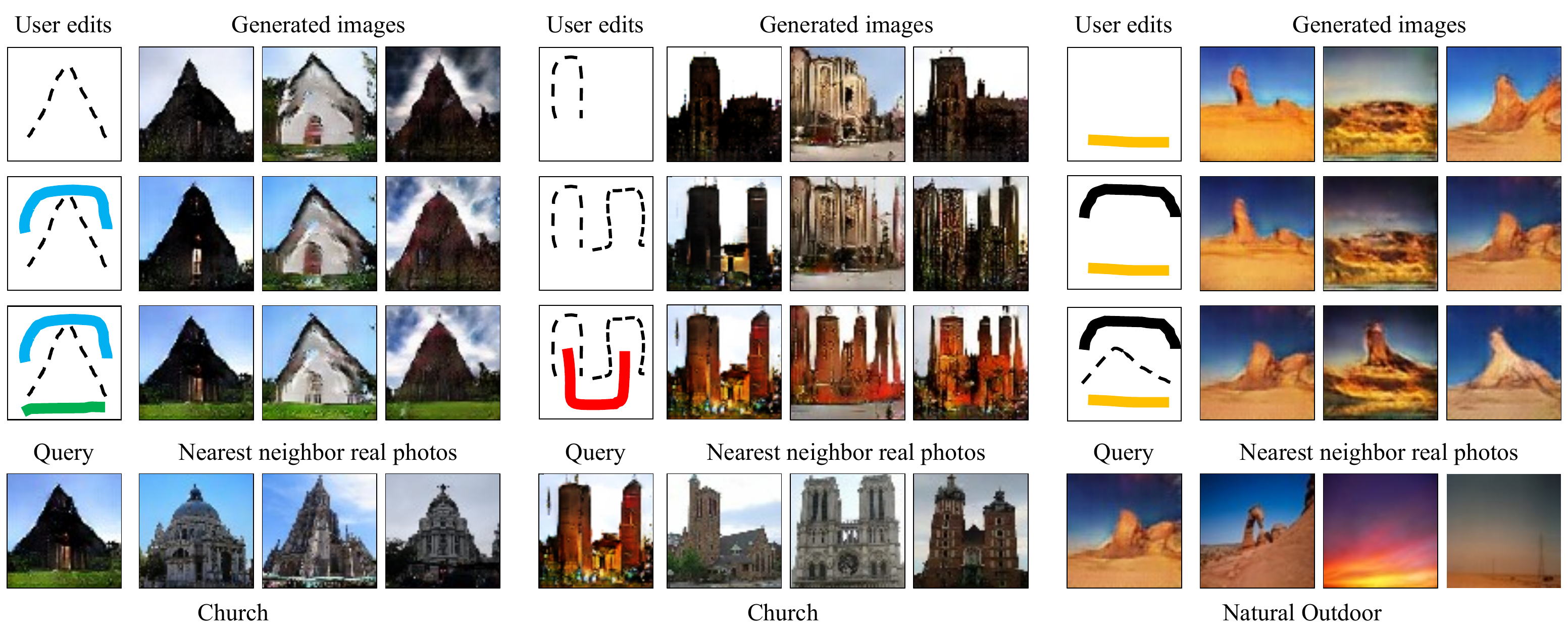}
\end{center}
   \caption{Interactive image generation. The user uses the brush tools to generate an image from scratch (top row) and then keeps adding more scribbles to refine the result (2nd and 3rd rows). In the last row, we show the most similar real images to the generated images. (dashed line for the sketch tool, and color scribble for the color brush)}
\label{fig:sketch2gen}
\end{figure}

Another byproduct of our method is that if there is no image to begin with and all we have are the user brush strokes, the method would generate a natural image that best satisfies the user constraints. This could be useful for dataset exploration and browsing. The difference with previous sketch-to-image retrieval methods~\cite{sun2013sketch} or AverageExplorer~\cite{zhu2014averageExplorer} is that due to potentially contradicting user constraints, the result may look very different than any single image from the dataset or an average of such images, and more of a realistic hybrid image~\cite{risser2010hybrids}. See some examples in \reffig{sketch2gen}.

\subsection{Evaluation}
\label{sec:eva}
\begin{table}
\begin{center}
 \begin{tabular}{l | c |  c | c | c|c}
  & Shoes  & Church Outdoor & Outdoor Natural & Handbags & Shirts   \\
  \hline
 \textit{Optimization-based}  &  0.155 & 0.319 & 0.176  & 0.299 & 0.284 \\
  \hline
  \textit{Network-based} & 0.210  & 0.338 & 0.198 & 0.302& 0.265 \\
   \hline
 \textit{Hybrid (ours)} & {\bf 0.140} & {\bf 0.250} & {\bf 0.145} & {\bf 0.242} & {\bf 0.184}   \\
\end{tabular}
\end{center}
 \caption{Average per-dataset image reconstruction error measured by $\mathcal{L}(x, x^R)$.}
\label{tbl:rec}
\end{table}

\vspace{2pt}\noindent {\bf Image reconstruction evaluation:} We evaluate three image reconstruction methods described in \refsec{project}: optimization-based, network-based and our hybrid approach that combines the last two. We run these on $500$ test images per category, and evaluate them by the reconstruction error $\mathcal{L}(x, x^R)$ defined in \refeq{rec_M}. ~\reftbl{rec} shows the mean reconstruction error of these three methods on $5$ different datasets. We can see the optimization-based and neural network-based methods perform comparably, where their combination yields better results. See~\reffig{rec} for a qualitative comparison. We include PSNR (in dB) results in the supplementary material.

\vspace{2pt}\noindent {\bf Class-specific model:}
So far, we have trained the generative model on a particular class of images. As a comparison, we train a cross-class model on three datasets altogether (i.e. shoes, handbags, and shirts), and observe that the model achieves worse reconstruction error compared to class-specific models (by $\sim10\%$). We also have tried to use a class-specific model to reconstruct images from a different class. The mean cross-category reconstruction errors are much worse: shoe model used for shoes: $0.140$ vs. shoe model for handbags: $0.398$, and for shirts: $0.451$. However, we expect a model trained on many categories (e.g. $1,000$) to generalize better to novel objects.

\vspace{2pt}\noindent {\bf Perception study:} We perform a small perception study to compare the photorealism of four types of images: real photos, generated samples produced by GAN, our method (shape only), and our method (shape+color). We collect $20$ annotations for $400$ images by asking Amazon Mechanical Turk workers if the images look realistic or not. Real photos: $91.5\%$, DCGAN: $14.3\%$, ours (shape+color): $25.9\%$; ours (shape only): $48.7\%$. DCGAN model alone produces less photo-realistic images, but when combined with our edit transfer, the realism significantly improves.

%% file: conclusion.tex
\section{Discussion and Limitations}

We presented a step towards image editing with a direct constraint to stay close to the manifold of real images. We approximate this manifold using the state-of-the-art in deep generative models (DCGAN). We show how to make interactive edits to the generated images and transfer the resulting changes in shape and color back to the original image. Thus, the quality of the generated results (low resolution, missing texture and details) and the types of data that DCGAN applies to (works well on structured datasets such as product images and worse on more general imagery), limits how far we can get with this editing approach. However, our method is not tied to a particular generative model and will improve with the advancement of this field. Our current editing brush tools allow rough changes in color and shape but not texture and more complex structure changes. We leave these for future work.

\vspace{4pt}\noindent  {\bf Acknowledgments}
This work was supported, in part, by funding from Adobe, eBay, and Intel, as well as a hardware grant from NVIDIA.
J.-Y. Zhu is supported by Facebook Graduate Fellowship.